# Development of digitally obtainable 10-year risk scores for depression and anxiety in the general population

**Morelli, D.[1,2], Dolezalova, N.[1], Ponzo, S.[1], Colombo M[1]. and Plans, D.[1, 3, 4]**

[1] Huma Therapeutics Ltd, London, United Kingdom

[2] Department of Engineering Science, Institute of Biomedical Engineering, University of Oxford, Oxford, United Kingdom

[3] Department of Experimental Psychology, University of Oxford, Oxford, United Kingdom
[4] INDEX Group, Department of Science, Innovation, Technology, and Entrepreneurship, University of Exeter, United Kingdom

**\* Correspondence:**
David Plans, PhD
Department of Experimental Psychology
University of Oxford
Woodstock Rd,
Oxford
OX2 6GG
United Kingdom
email: david.plans@psy.ox.ac.uk
phone: +44 (0) 7527016574

**Abstract**
The burden of depression and anxiety in the world is rising. Identification of individuals at increased risk of developing these conditions would help to target them for prevention and ultimately reduce the healthcare burden. We developed a 10-year predictive algorithm for depression and anxiety using the full cohort of over 400,000 UK Biobank (UKB) participants without pre-existing depression or anxiety using digitally obtainable information. From the initial 204 variables selected from UKB, processed into > 520 features, iterative backward elimination using Cox proportional hazards model was performed to select predictors which account for the majority of its predictive capability. Baseline and reduced models were then trained for depression and anxiety using both Cox and DeepSurv, a deep neural network approach to survival analysis. The baseline Cox model achieved concordance of 0.813 and 0.778 on the validation dataset for depression and anxiety, respectively. For the DeepSurv model, respective concordance indices were 0.805 and 0.774. After feature selection, the depression model contained 43 predictors and the concordance index was 0.801 for both Cox and DeepSurv. The reduced anxiety model, with 27 predictors, achieved concordance of 0.770 in both models. The final models showed good discrimination and calibration in the test datasets.We developed predictive risk scores with high discrimination for depression and anxiety using the UKB cohort, incorporating predictors which are easily obtainable via smartphone. If deployed in a digital solution, it would allow individuals to track their risk, as well as provide some pointers to how to decrease it through lifestyle changes.

# 1  Introduction

Global prevalence of depression was estimated to be 280 million[1] in 2019. By 2030, depression is expected to be the second-largest contributor to worldwide loss of years of healthy life because of death or disability[2]. Highly comorbid with depression, anxiety disorders globally are estimated to affect 301 million individuals[1]. NICE guidelines currently recommend the use of validated questionnaires (e.g. PHQ-9, Patient Health Questionnaire[3]; HADS, Hospital Anxiety and Depression Scale[4] and BDI, Beck Depression Inventory [5] for depression; GAD-2 or GAD-7[6] for anxiety disorders) to diagnose patients and classify the severity of their symptoms [7]. Whilst tools based on patients' self-reported feelings and mood changes are invaluable to track the progression of the disorder, multifactorial models including well-established risk factors are needed to successfully manage the disorder in the long-term. Predictive scores can be used to effectively identify patients at highest risk of developing depression or anxiety and enrol them in preventative pathways, thus minimising relapses and lowering the burden of the disease[8].

A comprehensive review of existing predictive scores is outside of the scope of this paper and can be found elsewhere[9]. There are several scores for specific populations at risk of depression, e.g. adolescents[10], elderly[11], traumatic head injury[12] or stroke patients[13], patients with diabetes[14] or immune-mediated inflammatory disorders[15]. For the general population, the most widely used depression risk score is the PredictD score, developed using patient data from six European countries and externally validated on a population from Chile[16]. The original score contains 10 risk factors (age, sex, country, education level, personal and family history of depression, physical and mental health disturbances, difficulties at work and experience of discrimination). Other country-specific scores have been developed after PredictD to better account for cultural and socio-economic differences [17–20], but little research has been conducted to develop risk scores aimed at predicting onset of generalised anxiety and panic disorders in the general population. The PredictA score was developed using the same dataset described above for identifying factors predicting depression, and it includes sex, age, lifetime depression, family history of psychological difficulties, physical health and mental health disturbances, unsupported difficulties in paid and/or unpaid work, country of residence and time of follow-up[21]. Given the high comorbidity between anxiety and depressive disorders, it has also been suggested that, on top of disorder-specific risk factors, a set of common underlying risk factors for both disorders may exist[22].

The majority of the published risk scores for depression and anxiety relate to short term predictions (between 6 and 24 months) that mainly involve non-modifiable factors (e.g. family history). Given the impact of recurring episodes and lifetime duration on the progression of the disorders [23,24], early identification of individuals at risk of depression or anxiety would be beneficial in devising effective preventative and therapeutic pathways. Evidence suggests that the risk of depression and anxiety can be decreased by modifying certain lifestyle factors, such as having a balanced diet and performing physical activity[25,26] or smoking cessation[27]. With an increased availability and need for telematic solutions in healthcare due to the COVID-19 pandemic[28–30], risk scores evidencing modifiable lifestyle changes have the potential to be of wide benefit.

The UK Biobank (UKB) is a prospective study of over half a million UK participants, recruited between 2006 and 2010. Available data includes primary care and hospital inpatient records, results of touchscreen questionnaires and verbal interviews about lifestyle, pre-existing conditions and family history, as well as a comprehensive battery of medical tests, imaging and physical assessments. While predictive models have been developed for many common diseases, such as cardiovascular diseases or diabetes [31,32], no long-term predictive models have been derived for depression and anxiety using the UKB cohort. This dataset contains information about many known predictors of severe mental illness, including modifiable lifestyle factors and other digitally-obtainable data such as comorbidities, socioeconomic factors or early-life events [33,34].

The aim of this study is to devise a model of potential risk factors for depression and anxiety in the long term (10+ years) using the UKB data and with a focus on behavioral and health indices that can be potentially tracked by means of a remote digital solution.

## 2 Methods

### 2.1 Data source and study design

The use of data for this study was approved by UKB, under the project title "Validation and comparative analysis of novel prediction models focused on modifiable lifestyle factors for the risks of common, preventable diseases and all-cause mortality: a cohort study" (application number 55668).

The aim of this study was to model the risk of being diagnosed with depression or anxiety for the first time over the next 10 years. The outcomes were defined as an occurrence of a depressive episode (ICD10 code F32) or an anxiety disorder (ICD10 code F41), respectively, after the date of assessment. These were derived from the UKB First Occurrences fields 130894 and 130906 which combine data from primary care and hospital inpatient diagnoses with conditions self-reported at the time of assessment.

Potential predictor variables include results from the touchscreen questionnaire administered at the initial assessment (instance 0) and the online mental health questionnaire introduced in 2009, as well as pre-existing illnesses diagnosed by the time of assessment. Only fields which could be collected via a smartphone app (i.e., using the sensors available on smartphones or via user's input) were included. The final set of potential predictors includes demographic data (age, sex, ethnicity), socioeconomic status (income, qualifications), physio-metric data (body mass index, height, weight, etc.), family history (illnesses of parents or siblings), medical history, lifestyle characteristics (physical activity, diet, sleep habits), traumatic events in early life and adulthood, addictions, moods and overall perceived wellbeing (satisfaction with life, mood swings, feelings of worry, loneliness, etc.). Fields with a high proportion of missing values were excluded, as well as fields which could cause label leakage (e.g., 'Seen a psychiatrist for nerves, anxiety, tension or depression').

### 2.2 Data preparation

Participants were excluded from the study if they were diagnosed with the corresponding outcome condition prior to the date of assessment. Survival time was measured as years from the date of assessment (instance 0 in the UK BioBank dataset) until the date of depression/anxiety diagnosis, or in participants who were not diagnosed, data was right-censored at the data extraction date (30th September 2020), date of death or date when they were lost to follow-up.

The binary variable "any_mental_issue" was derived from the set of First Occurrences fields for ICD10 diagnoses in the 'Mental and behavioral disorders' category (F00–F72, excluding F32 and F33 in the Depression model and F40 and F41 in the Anxiety model). Only dates prior to the date of assessment were considered for this variable.

All categorical features were one-hot encoded, followed by exclusion of categories containing less than 0.1% items. Continuous features were centered and scaled to unit variance. Finally, participants with any remaining missing data were excluded.

Test set (25 %) was set aside for internal validation using a stratified train-test split (preserving the ratios seen in the binary outcome field). From the train set, a further 25 % was set aside as a validation set and used for the optimisation of DeepSurv parameters. For the final models after feature selection, the train and validation dataset were combined for training, followed by evaluation on the test dataset.

### 2.3 Baseline model and feature selection

First, a Cox-Proportional Hazards (CPH) model, implemented in the Python lifelines library[35], was trained using the full set of features. The number of features was then reduced to decrease the risk of overfitting, improve the explainability of the model and to narrow down the number of inputs from users in a potential digital solution. As a first step, univariate Cox analysis was performed for each feature and those with p-value over 0.1 were excluded. Then, an iterative backward elimination algorithm was used to get the final set of features. In short, in every round of elimination, the CPH model was retrained without a set of features with the highest p-value. If the concordance index evaluated on the validation dataset decreased by more than 0.001, features were kept for an additional round testing elimination of a smaller number of features. Elimination of each remaining single feature was tested before the decision to keep it in the reduced model.

The reduced model was then trained using this final set of features and its performance evaluated on the unseen test cohort.

### 2.4 Deep Survival Analysis

The next model we tested was the Cox proportional hazards deep neural network (DeepSurv), using an implementation in the 'pycox' package[36] based on the deep learning library PyTorch[37].

Using the full set of features, we searched the hyperparameter space using a set of parameters described in *Supplementary Table 1*. This was done using a Tree-Structured Parzen Estimator algorithm[38] from the Optuna Library[39]. In total, 200 configurations were tested for both depression and anxiety, each evaluated by 3-fold cross-validation. Feedforward neural networks deep up to 3 hidden layers have been tested. Classic Stochastic Gradient Descent algorithms with Momentum [40] and Adam[41] with optimal learning rate estimation were used for training. The best combination of hyperparameters, optimised on the full set of variables, was selected separately for the depression and anxiety models. The performance was then compared to a neural network with the same hyperparameters but using only the reduced set of features selected by the backwards elimination using Cox classifier.

### 2.5 Statistical analysis

Results from the analysis of demographic characteristics show participant numbers and percentages of total for categorical/ordinal variables, or medians and quartiles (Q1 and Q3) for continuous variables. Statistical comparisons were performed using the Chi-squared test for categorical/ordinal and Kruskal-Wallis test for continuous variables.

C-index was used as the metric for all models, with 95% confidence intervals calculated using the percentile bootstrap resampling method (50 resampling rounds). Where detailed analysis of the results of CPH models is provided, logarithm of hazard ratios/log(HR) with 95% confidence intervals (CIs) are shown. P-values test the null hypothesis that the coefficient of each variable is equal to zero and significance level was set to 0.05. Calibration was evaluated at the 10-year time point using calibration plots and the Integrated Calibration Index (ICI), which is a mean weighted difference between observed and predicted probabilities, implemented in the Python lifelines library[35].

## 3 Results

### 3.1 Study population

From the initial set of 502,488 participants in the UK Biobank, 63,927 had pre-existing depression and 31,876 pre-existing anxiety. These participants were excluded from the respective datasets. Further participants were excluded due to missing values in some continuous or ordinal variables. In the depression dataset, of the remaining 425,517 participants, 13,752 (3.23 %) developed depression after assessment. For the anxiety dataset, it was 15,534 (3.40 %) out of 456,815 participants diagnosed with anxiety after assessment. Details on distribution of the outcomes in the train, validation and test datasets can be found in *Supplementary Figure 1*. There were no significant differences in any features between the train, validation and test datasets (data not shown). Median follow-up time of 11.6 years and maximum follow-up time of 13.8 years were the same for depression and anxiety.

The dataset used in this study contains 45.6% men and 54.4 % women, aged 56.5 ± 8.1 (range 37–73) at the time of the initial assessment. The ethnic background of the participants was 94.1 % white, 2.3 % Asian, 1.6 % black and 1.5 % other. Summary of the variables in the final model for depression and anxiety is presented in *Supplementary Tables 2 and 3,* respectively.

**3.2  Cox model and feature selection**

The initially selected 204 UK Biobank variables were pre-processed into 523/529 feature columns for depression/anxiety. These were used to build a baseline Cox proportional hazards model for each of the two outcomes, achieving a concordance index of 0.813 for depression and 0.778 for anxiety in the validation cohort (*Table 1*).

After feature selection, the number of predictors for depression was narrowed down to 43, accompanied by a slight decrease of the concordance index to 0.801 for test dataset (*Table 1*). For anxiety, the final number of predictors was 27, achieving concordance of 0.770 for test dataset (*Table 1*). The predictors in the final models, along with their coefficients and confidence intervals are shown in *Figure 1* and *Supplementary Tables 4 and 5*. All predictor coefficients in the final models were statistically significant at the 0.05 level.

|  | **Before feature selection** | | | **After feature selection** | | |
| --- | --- | --- | --- | --- | --- | --- |
|  | # features | C-index (train) | C-index (validation) | # features | C-index (train + validation) | C-index (test) |
| Depression | 523 | 0.8100 [0.8091, 0.8108] | 0.8125 [0.8109, 0.8144] | 43 | 0.8071 [0.8068, 0.8074] | 0.8009 [0.8004, 0.8015] |
| Anxiety | 529 | 0.7822 [0.7810, 0.7832] | 0.7780 [0.7760, 0.7799] | 27 | 0.7764 [0.7761, 0.7765] | 0.7699 [0.7695, 0.7703] |

*Table 1: Results of the Cox Proportional Hazards model for depression and anxiety. Mean concordance index is shown for training and validation/test dataset. 95 % confidence intervals are shown in square brackets.*

The top three risk factors in the depression model were being diagnosed with mania, hypomania, bipolar or manic-depression by a professional, being diagnosed with anxiety by a professional and feeling down, depressed or hopeless nearly every day in the past two weeks. The most protective factors were never visiting a GP for nerves, anxiety, tension or depression and not avoiding (or only a little bit avoiding) situations which remind them of a stressful experience. For anxiety, the top risk factors being diagnosed with panic attacks or depression by a professional and suffering from nerves, the protective factors were never visiting a GP for nerves, anxiety, tension or depression, never being diagnosed with life-threatening illness and excellent self-reported health rating.

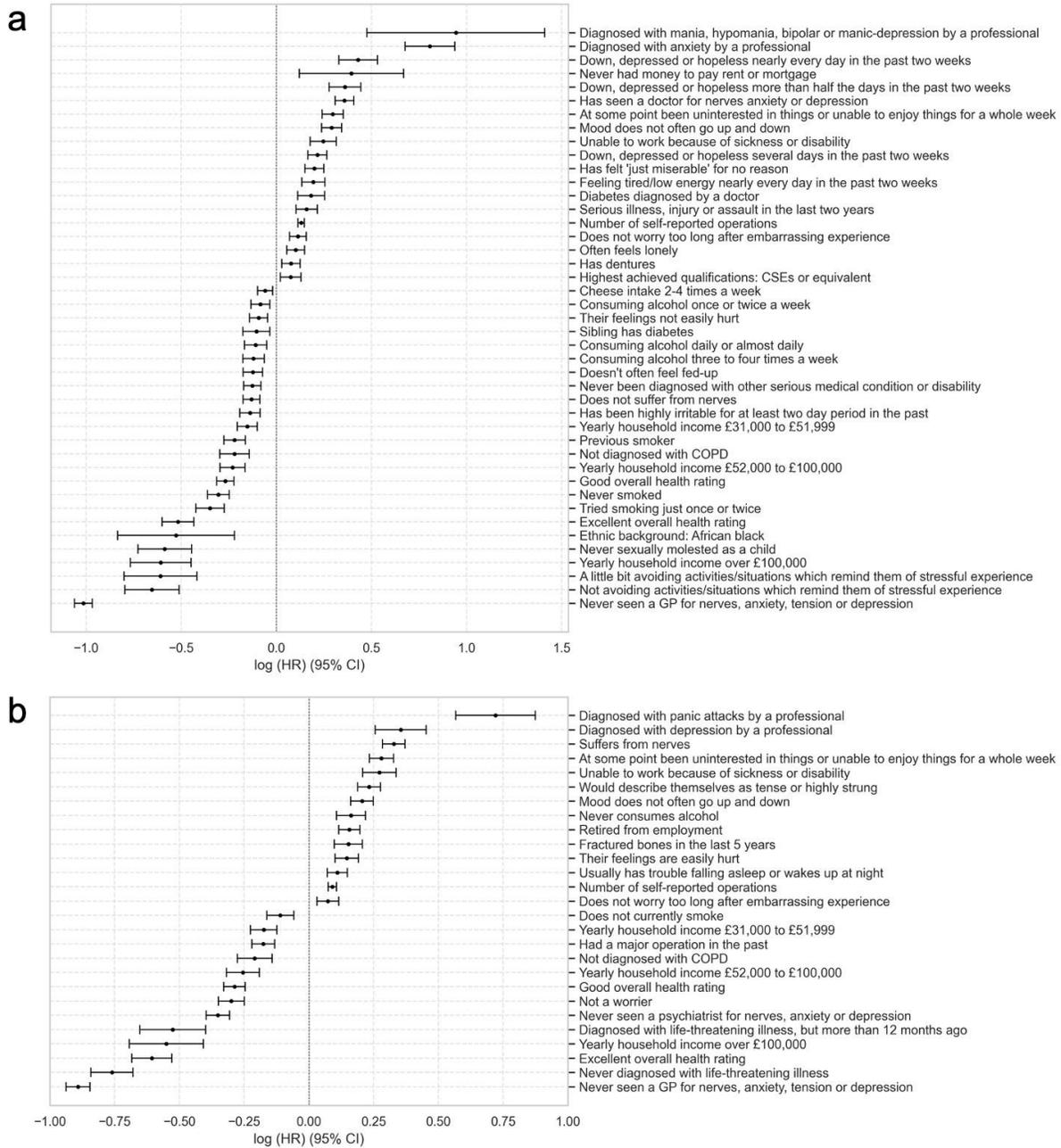

*Figure 1: Plot of Cox Proportional Hazards model coefficients for depression (a) and anxiety (b). Values show log(HR) ± 95% CI. HR = hazard ratio, CI = confidence interval.*

The mean predicted 10-year risk of developing depression was 2.90 % (95% CI 2.88–2.93), the mean observed risk was 2.83 % (95% CI 2.80–2.85). The anxiety model predicted an average risk of 2.94 % (95% CI 2.92–2.96), while the observed probability was 2.90 % (95% CI 2.88–2.92). The Cox models showed good calibration, particularly for the low probabilities which were abundant in the population, with slightly larger errors for the higher probabilities which were sparsely represented. The Integrated Calibration Index (ICI) of 0.085 % for depression and 0.11 % for anxiety (***Supplementary Figure 2***).

### 3.3 Machine learning models

The hyperparameter space for the DeepSurv model was explored by running 500 combinations, best of which achieved a concordance index of 0.805 (95 % CI 0.804–0.806) for depression and 0.774 (95% CI 0.772–0.775) for anxiety using the full set of variables (*Table 2*). Using the reduced set of features after feature selection, the depression model showed a concordance of 0.801 (95% CI 0.800–0.802) and the anxiety model concordance of 0.770 (95% CI 0.769–0.771) (*Table 2*). The details of the hyperparameters for the best-performing models can be found in *Supplementary Table 1*.

*Table 2: Summary of the best-performing DeepSurv models for depression and anxiety. Trial number is shown along with the hyperparameters of the best model and mean C-index calculated on the test dataset, before and after feature selection. 95% confidence intervals are shown in square brackets.*

|  | **Before feature selection** | | **After feature selection** | |
| --- | --- | --- | --- | --- |
|  | # Features | C-index | # Features | C-index |
| **Depression** | 523 | 0.8051 [0.8041, 0.8061] | 43 | 0.8011 [0.8002, 0.8020] |
| **Anxiety** | 529 | 0.7737 [0.7724, 0.7751] | 27 | 0.7702 [0.7692, 0.7709] |

### 3.4 Comparison of the depression and anxiety models

Of the 43 features in the depression model and 27 features in the anxiety model, 12 common features appeared in both models (**Figure 2**). These include having been uninterested in things for at least a week in the past, being moody, being unable to work because of sickness or disability, increased number of operations and not worrying too long about an embarrassing experience. Conversely, an annual household income of £31,000 to £100,000, not being diagnosed with COPD, good or excellent self-reported overall health-rating and never seeing a GP for nerves, anxiety, tension or depression had negative coefficients in both models.

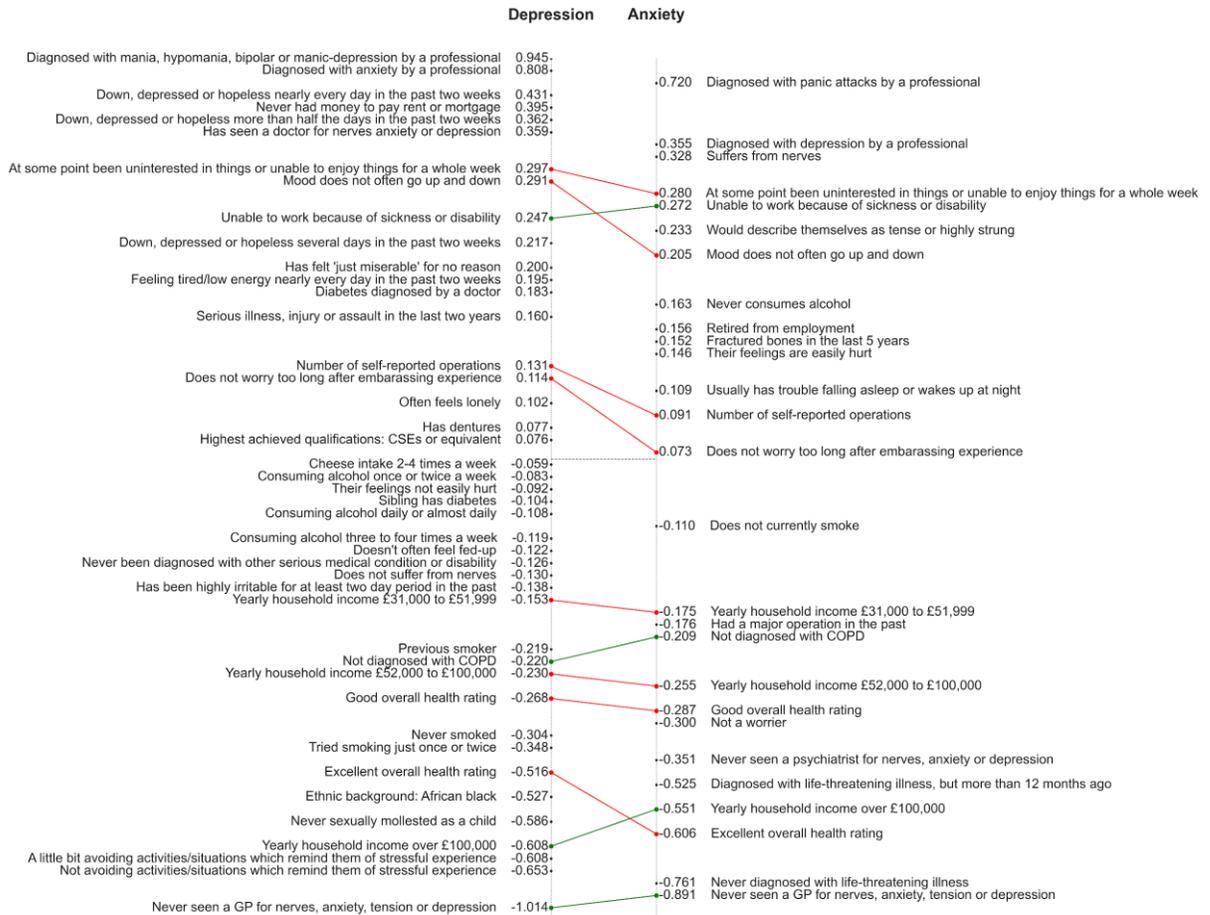

*Figure 2: Visual comparison of the reduced models for depression and anxiety.* Features are positioned on the y axes in the order of descending coefficients but the distance between points is not proportional to the difference between coefficients. Connecting lines are shown in red (for common factors with higher coefficient in the depression model) or green (higher coefficient in anxiety model).

## 4 Discussion

### Key findings

The aim of the current study was to build prediction models for depression and anxiety, with a specific focus on factors linked to digitally-obtainable data. Using a data-driven approach to feature selection and model optimisation, we trained models for prediction of depression and anxiety using both traditional statistical and machine learning methods. The best-performing model for depression achieved a concordance index of 0.801, slightly overperforming other similar depression risk scores. For comparison, the current golden standard score PredictD achieved a concordance index of 0.790 using stepwise logistic regression[16]. Rosellini *et al.* used an ensemble machine learning algorithm with a concordance of 0.757[19], whereas Wang *et al.* developed sex-specific logistic regression models with concordance index of 0.795 for men and 0.767 for women[20]. Our DeepSurv model for anxiety shows a concordance of 0.770, compared to 0.752 in the PredictA study[21]. It is important to note that participants in our study were followed for more than ten years while the prediction horizon in the other studies was shorter (1–4 years). Among the most similar studies developed on the UKB dataset, Zhou *et al.* achieved a concordance index of 0.778 for prediction of depressive moods using neuroimaging and questionnaire data and Sarris *et al.* in their study of lifestyle factors associated with frequency of depressive moods built an ordinal logistic regression model but did not report its performance[42]. To the best of our knowledge, this is the first study which used the UKB cohort to develop long-term prognostic risk scores for both depression and anxiety.

The DeepSurv model performed comparably to the Cox model. While it has the capability to capture complex non-linear relationships between factors, this fact could point to a linear relationship of most variables included in the model. For the use of this score in a digital healthcare setting, interpretability of the score is key. With DeepSurv, similar to many other black-box machine learning models, the direction and scale of each feature's contribution to the overall risk may not be easily understandable. From this perspective, Cox model coefficients provide more intuitive understanding of how each feature could be changed to decrease the predicted risk, an information which could potentially motivate the user towards the right lifestyle changes.

Our prediction model includes many traditional risk factors for mental health illness, including smoking, alcohol consumption, pre-existing diabetes or COPD, employment status, overall health status, sleeping disturbances or education level [33]. Interestingly, while age and sex are included as risk factors in the existing predictive scores[16,18,21], they were eliminated from our models during the feature selection. While this data-driven approach eliminated some of these traditional risk factors, the resulting set of predictors is capable of slightly better discrimination than the existing scores where predictor selection was based mainly on evidence in literature.

From the comparison of the final set of predictors in the depression and anxiety models it is clear that some pre-existing mental health conditions adversely affect the risk: being diagnosed with anxiety increases the risk of having depression in the future and vice versa. In addition, risk of depression is increased with pre-existing diagnosis of mania, hypomania or bipolar disorder, whereas the risk of anxiety by previous diagnosis of panic attacks. Reported period of being uninterested in things for at least a week in the past features as a risk factor in both models. Conversely, never seeing a GP for nerves, anxiety, tension or depression has the highest negative coefficient in the model. Other factors featuring in both depression and anxiety models include the self-reported overall health rating, as well as associated variables of number of operations or inability to work because of sickness or disability. Our analysis also showed that household income or certain personality traits (being a worrisome or moody person) are common risk factors for depression and anxiety.

Some of the predictors in our final model are dynamic factors, changing over time (e.g. smoking status, alcohol consumption, down, depressed or hopeless feelings or tiredness over the past 2-week period). Regular reassessment of the score would therefore be beneficial and could be aided by deployment of the score as a digital solution. Such application could also provide personalised recommendations on how to change one's lifestyle to decrease the risk of development of depression or anxiety. The small number of lifestyle factors

### Study limitations

Among the limitations of this study is the selection bias in the UKB cohort used to train the prediction models. It has been reported that the participants of UKB are on average healthier and come from less deprived areas[43]. Importantly, the representation of ethnicities in the dataset is deviating from the general population, with over 94 % participants being white, compared to 86 % in the general population according to the latest UK census. Caution should therefore be exercised when making predictions for individuals with a minority background, as they were underrepresented in the training dataset. This is an important limitation because they may be disproportionately affected by severe mental illnesses[44]. The study also included only participants with a limited age range: 37–73 years at the time of assessment. Therefore, until these models have been validated on an appropriate cohort extending this range, predictions for users younger or older than this should be interpreted with reserve. The results of the internal validation performed in this study using an unseen test cohort does not point to overfitting but external validation will be necessary to confirm this.
Missing records might also be introducing potential bias into this study. Several important variables from UKB came from the mental health questionnaire, completed by only about 150,000 participants. Using one-hot encoding, we were able to utilise these variables and increase the predictive accuracy for participants who had this information available. Our assumption was that this data was missing at random but further research is needed to assess its potential role in biasing the data.

### Implications of our findings

This tool is intended to be used as a digital solution for dynamic tracking of individuals' risk of developing depression or anxiety. Evidence suggests that most individuals value knowing their risk of depression, especially if ways of prevention are also indicated [45,46]. Providing a personalised depression/anxiety risk score therefore appears as a safe preventative strategy with demonstrated benefits[46].

External validation of the developed model on data collected from another population will be necessary to be able to extend these findings to populations from geographies other than the United Kingdom. If these populations show significant differences in demographic characteristics, such as ethnicity, causing the accuracy of the predictions to drop, the model can be re-calibrated if a suitable dataset exists for this population.

Additional improvements to the prediction models for mental health would be an expansion of the input space with other dynamic parameters, such as activity measures. Raw accelerometer data is available for a proportion of UK Biobank participants, and it could be used to provide an objective view on individuals' level of activity, sedentary time and sleep, with known associations with the risk of development of mental illness [33,42,47]. Because these would be obtainable via wearable devices, they bring the potential of truly dynamic monitoring of an individual's risk.

### Conclusion

In summary, we present algorithms for prediction of depression and anxiety, developed using a large UK cohort of individuals followed for over 10 years. All factors in our models are easily acquirable via smartphone devices and thus can be used to support development of preventative digital solutions for mental health.

**Data Availability Statement**
The dataset analyzed in this study has been provided by the UK BioBank.

**Author Contributions**
DM designed the study, MC assisted with implementation of the used methods, DM and ND trained the models and performed data analysis, DM, ND, SP and DP wrote the manuscript, all authors revised the manuscript.

**Conflict of Interest**
DM, ND, SP, MC and DP are current employees of Huma Therapeutics LTD.

**10y Risk Scores For Depression And Anxiety**

**10y Risk Scores For Depression And Anxiety**